\documentclass[11pt,a4paper]{article}
\usepackage[hyperref]{emnlp2018}
\usepackage{times}
\aclfinalcopy
\usepackage{latexsym}
\usepackage{amsmath, amssymb, amsfonts, amsthm}
\usepackage{graphicx}
\usepackage{url} 
\usepackage{bbm}
\usepackage{color}
\usepackage{bm}

\usepackage{subfigure}
\usepackage{mathabx}
\usepackage{floatrow}
\usepackage{wrapfig,lipsum,booktabs}
\usepackage{multirow}
\usepackage{hhline}
\definecolor{mygreen}{rgb}{0.35, 0.50, 0.30}

\def\aclpaperid{388} 

\usepackage{mathtools}
\newcommand{\beq}{\vspace{0mm}\begin{equation}}
\newcommand{\eeq}{\vspace{0mm}\end{equation}}
\newcommand{\beqs}{\vspace{0mm}\begin{eqnarray}}
\newcommand{\eeqs}{\vspace{0mm}\end{eqnarray}}
\newcommand{\barr}{\begin{array}}
\newcommand{\earr}{\end{array}}

\newcommand{\Lcal}{\mathcal{L}}

\newcommand{\Tcal}{\mathcal{T}}
\newcommand{\Pcal}{\mathcal{P}}

\newcommand{\ie}{\textit{i.e.}}

\title{Consistent Dialogue Generation with Self-supervised Feature Learning}

\author{Yizhe Zhang \quad\quad Xiang Gao \quad\quad Sungjin Lee \\ \textbf{Chris Brockett}\quad\quad \textbf{Michel Galley}\quad\quad \textbf{Jianfeng Gao}\quad\quad \textbf{Bill Dolan}\\
  Microsoft Research, Redmond, WA, USA \\
  {\small \tt \{yizzhang,xiag,sule,chrisbkt,mgalley,jfgao,billdol\}@microsoft.com}
}

\date{}

\begin{document}
\maketitle
\begin{abstract}

Generating responses that are consistent with the dialogue context is one of the central challenges in building engaging conversational agents. 
In this paper, we propose a neural conversation model that generates consistent responses by maintaining certain features related to topics and personas throughout the conversation.
Unlike past work that requires external supervision such as user identities, which are often unavailable or classified as sensitive information, our approach trains topic and persona feature extractors in a self-supervised way by utilizing the natural structure of dialogue data. Moreover, we adopt a binary feature representation and introduce a feature disentangling loss which, paired with controllable response generation techniques, allows us to promote or demote certain learned topics and personas features. 
The evaluation result demonstrates the model's capability of capturing meaningful topics and personas features, and the incorporation of the learned features brings significant improvement in terms of the quality of generated responses on two datasets, even comparing with model which explicit persona information. 
\end{abstract}

\section{Introduction}

The notion of speaker consistency is attracting growing interest in neural response generation research \cite{li2016persona,luan2016multiplicative,zhang2018personalizing,gaosurvey}. When interacting with an open-domain neural conversation agent, users may expect the agent to develop the dialogue with consistent information, mitigating the user confusion and improving engagement. 
Speaker consistency presents two aspects: \textit{topic consistency} and \textit{persona consistency}. Topic consistency reflects the model's ability to maintain dialogue topics such as sport, movie or music without getting sidetracked. Persona consistency envisions the agent as human-like, endowed with a relatively invariant individual personality,  style of engagement (e.g., enthusiasm and casualness) or personal profile (e.g., place of residence). 

\begin{figure}[ht!]
    \centering
    \includegraphics[width=.8\textwidth]{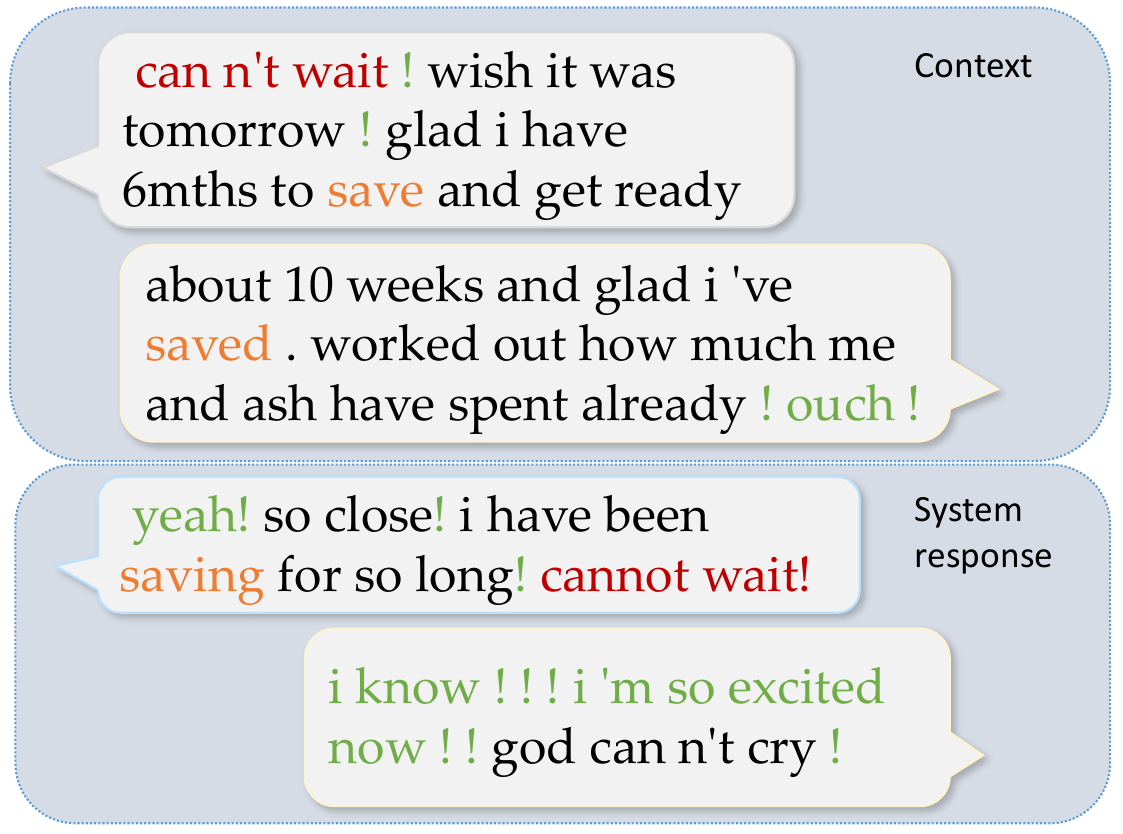}
    \caption{Task illustration: generating responses that are consistent with dialogue history in \textcolor{red}{persona}, \textcolor{mygreen}{tone} and \textcolor{orange}{topic} (from our system, 2 context turns).} 
    \label{fig:task}
\end{figure}

Generating appropriate responses with these characteristics is a major challenge (Figure~\ref{fig:task}).
\citet{li2016persona, luan2017multi} and \citet{al2016conversational} use persona embeddings as the additional input to train end-to-end conversational agents. Obtaining accurate persona embeddings as in \citet{li2016persona} however requires many thousands of utterances per persona, 
and targeted test personas may not always be found in the training data.
End-to-end systems are often trained from social media data in which only a small spectrum of personas (casual speakers) is represented and professional roles (e.g. customer service) may be underrepresented, thus limiting deployment. 
Typically, moreover, the objective is to maintain consistency of both persona and topic throughout the dialogue, rather than inject specific personas/topics in responses.
Under these scenarios, learning and leveraging persona or dialogue topic in a data-efficient and unsupervised way becomes crucial. 



We present a self-supervised approach that uses the natural structure of conversational data to learn and leverage topic and persona features. Our proposals include:


1) A discriminative feature extraction mechanism that captures conversational topics and personas in a self-supervised manner, without requiring \textit{specification of speaker identity}, thus allowing massive unlabeled datasets to be utilized while protecting sensitive user information. 

2) Use of binary features and a disentangling loss to improve interpretability of learned features. This affords flexibility to activate or deactivate specific features when generating responses.   

3) Leveraging a controllable text generation mechanism to force generated responses to adhere to high-level features such as topic and persona encoded in the controlling signal.



\section{Related Work}


\paragraph{Self-supervised learning} Self-supervised 
as a subdomain of unsupervised learning, has been applied to representation learning for image, video and audio \cite{denton2017unsupervised,doersch2015unsupervised,owens2018audio}. However, to the best of the authors' knowledge, the application of self-supervision in conversational agents is rare. Borrowing definitions from other domains, self-supervised approaches in NLP make use of non-textual signals that intrinsically correlate with the text to supervise the text feature learning \cite{denton2017unsupervised}.


\paragraph{Persona-aware response generation}
\citet{welleck2018dialogue} suggested a natural language inference (NLI) approach to improve the persona consistency, however additional labels are required. 
\citet{zhang2018personalizing, qian2017assigning} use explicit personal profiles as side information to guide response generation. Such information, however, may not always be available. Other work proposes injecting either emotion \cite{zhou2017emotional} or functional control \cite{ke2018generating} into dialogue generation. As in \citet{li2016persona}, learning to leverage the controlling signal in order to bias generation may require significant amounts of labelled data.

\paragraph{Topic-aware response generation} Leveraging topic modeling in response generation has been explored by several prior works \cite{xing2017topic, wang2017steering, wu2018response}. Our approach differs from these methods in that we focus on learning discriminative features that help distinguish a topic or person from another. Also, our method employs a neural sentence encoder to capture richer features than the bag-of-words features that the conventional topic models opt for. 

\paragraph{Interpretable and controllable generation}  Controllable text generation~\cite{hu2017toward} has been employed in text style transfer and many other tasks \cite{ficler2017controlling, asghar2018affective, ghosh2017affect, dong2017learning}. This helps disentangling high-level style information from contextual information such that the style information can be independently manipulated to produce text with different styles.
Related to our work, \citet{zhao2018unsupervised} considered discrete latent actions to learn a human-interpretable representation for task-oriented dialogue systems.




\section{Proposed Approach}
The proposed approach use additional unsupervisedly learned features to generate response utterances that reflect these features. We elaborate two major components of the proposed approach: a \textit{feature extractor} trained to extract topic/persona features from each utterance; and a \textit{response generator} that takes the extracted features as input to generate responses accordingly.
\subsection{Problem statement}
Let $d^{(i)} \triangleq [u_1^{(i)}, u_2^{(i)}, \cdots, u_T^{(i)}]$ denote the $i$-th dialogue session in dataset $\mathcal{D}$, where $u_j^{(i)}$ is the $j$-th utterance and $T$ is the number of turns in this dialogue session. 
We assume that each dialogue $d_i$ only consists of the utterances between two speakers, interleaving with each other. 
Suppose the first $K$ ($K<T$) turns of each dialogue are revealed, our aim is to generate the remaining $T-K$ turns of the dialogue that are consistent with the observed context.

\subsection{Discriminative feature extraction}
\label{sec:d}
\begin{figure}[ht!]
    \centering
    \includegraphics[width=.7\textwidth]{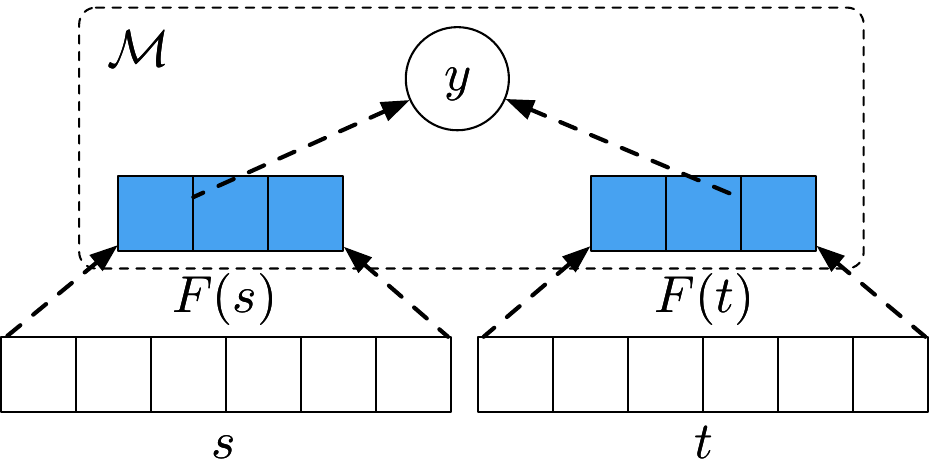}
    \caption{Feature extractor design. $s$ and $t$ are two randomly shuffled sentences. An extractor network $F(\cdot)$ ($F$ can be either $\Tcal$ or $\Pcal$) encodes both of them to yield features $F(s)$ and $F(t)$, which are then used to predict label $y$. $\mathcal{M}$ represents matching function/network}
    \label{fig:d}
\end{figure}
Inspired by \citet{denton2017unsupervised}, we adopt a self-supervised discriminative training scheme where we design a neural model which includes an explicit feature extraction layer as illustrated in Figure~\ref{fig:d} and formulate a discriminative task to train the model.
When training is done, the feature extraction layer yields relevant features for the associated task. 
In this section, we introduce two discriminative tasks to capture two types of sentence features, respectively: 1) \textit{topic-specific features} ($\Tcal$) that characterize conversation topics. 2) \textit{persona-specific features} ($\Pcal$) that reflect speaker characteristics.  

\paragraph{Topic feature extractor}
In order to build a topic feature extractor with self-supervision, we rely on the assumption that utterances from the same conversation session are likely to share similar topics. Thus, we formulate a surrogate task to identify if two random sentences $s$ and $t$ from $\mathcal{D}$ belong to the same dialogue session.
Specifically, when they come from the same dialogue session, \ie  $s,t \in d_i$, we assign 1 to target $y$ and 0 otherwise. 
We optimize the cross-entropy objective: 
\begin{align*}
    \mathcal{L_\texttt{xent}}& = \frac {1} {N}\sum_{i=1}^N [ y_i \log \mathcal{M}(\Tcal(s_i), \Tcal(t_i))  \nonumber \\
      & + (1-y_i) \log [1-\mathcal{M}(\Tcal(s_i), \Tcal(t_i))]],
\end{align*}
where $\Tcal(\cdot)$ denotes the topic-specific feature extractor (shared among all sentences), and $\mathcal{M}(\cdot,\cdot)$  represents a matching network detecting 
whether the two feature vectors $\Tcal(s_i)$ and $\Tcal(t_i)$ belong to the same dialogue session.
We use a 3-layer convolutional neural network (CNN) followed by a non-linear mapping for $\Tcal(\cdot)$ to produce an $L$-dimensional vector. For the non-linear mapping, we explore two options: 1) We employ a sigmoid function to produce a soft-binary representation, \ie $\Tcal(x) \in (0,1)^L$. 2) We compute a hard-binary representation by taking 1 if it is positive and 0 otherwise, \ie $\Tcal(x) \in \{0,1\}^L$. This non-negative bounded representation lends itself well to interpretation and control of each component of $\Tcal(x)$. For instance, we can activate or deactivate a certain topic or persona by simply turning on and off the corresponding component. 
For the matching function, we apply a sigmoid function to the inner-product of two feature vectors, \ie $f(\Tcal(s),\Tcal(t)) = \sigma (\Tcal(s)\cdot \Tcal(t)/\tau)$, where $\tau$ is a hyperparameter to scale the $\Tcal(s)\cdot \Tcal(t)$.

\paragraph{Persona feature extractor}Here we consider extracting persona features in a broader sense of current speaker's status related to emotion \cite{zhou2017emotional}, personality, tone and function control~\cite{ke2018generating}. Note that we are only interested in maintaining consistency of any emerging persona, rather than characterizing a full spectrum of persona features. The only difference between the topic ($\Tcal$) and persona ($\Pcal$) feature extractors is how the positive and negative sample pairs for training are created. In the persona feature extractor, the \textit{positive pairs} ($y=1$) or \textit{negative pairs} ($y=0$) are the utterances from the \textit{same} or \textit{different} speaker within a dialogue, aiming to eliminate the topic information from the persona features. Ideally, the two speakers in a dialogue are discussing the same topic. Under this assumption, since the utterances in a negative pair are also from the same dialogue, they are like to share the same topics. Thus, the model is forced to learn the features that can capture different personas of the two speakers. 

Unlike \citet{li2016persona}, where each speaker is assigned a single speaker embedding vector, in our proposed method the utterances by one speaker can have different feature vectors as the manifestations of the underlying persona embedding in a different context. Nevertheless, the discriminator objective encourages these vectors to be similar since they refer to the same person. We believe that our approach is more data-efficient than \cite{li2016persona} because the former allows borrowing information from a wider range of speakers. 
In \cite{li2016persona}, information borrowing only happens to the speakers who are similar to the current speaker in persona embedding space. As a result, persona embeddings can be poor for those not based on many dialogues.
Our method, on the other hand, can leverage those speakers who share any specific features with the current speaker and is able to learn more robust representations of speakers because we aggregate personal traits across all users. However, \citet{li2016persona} complement our methods nicely in that it does not require dialogue history as the seed to initiate the first several turns.  

\paragraph{Interpretable features}
We considered two methods of making the learned features more \textit{interpretable}: 1) feature vector disentanglement  \cite{cogswell2015reducing}; 2)  feature vector binarization \cite{zhao2018unsupervised}. 


First, we employ a \textit{decorrelation} (\texttt{DeCorr}) loss inspired by \citet{cogswell2015reducing}, who introduced a \texttt{DeCov} loss to regularize deep neural networks. Specifically, we add an additional term in the objective function when training the topic and persona feature extractors: 
\begin{align}
    &\mathcal{L}_\texttt{DeCorr}= \frac {1} {2} (||M||_F^2 - ||\texttt{diag}(M)||_F^2)  \nonumber \\
    & M_{jk}  = \frac{\sum_i (F(s_j) - \mu_j)(F(s_k) - \mu_k) } {\sqrt{\sum_i (F(s_j) - \mu_j)^2\sum_i(F(s_k) - \mu_k)^2}},\nonumber
\end{align}
where $||\cdot||_F$ represents the matrix Frobenius norm, and the $\texttt{diag}(\cdot)$ operator represents diagonalization of a matrix. $F$ denotes the feature extractor, and can be either $\Tcal$ or $\Pcal$. $M$ is the correlation matrix of $F$, computed from the current batch of data. Note that achieving a reasonable estimation of the correlation matrix requires a relatively large mini-batch size. The resulting final objective for the discriminator is $\mathcal{L}_\texttt{xent}+\lambda  \mathcal{L}_\texttt{DeCorr}$, where $\lambda$ is a balancing hyperparameter. 

Second, alternatively, we also consider \textit{binary} feature vectors, where a straight-through (ST) estimator is used for the gradient calculation \cite{bengio2013estimating, shen2018nash}. Suppose the binary feature $F$ is rounded from a probability vector $p$, ST estimator back-propagate through the hard threshold by approximating the gradient $\partial F/ \partial p$ as 1. 
We empirically found that setting $\mathcal{M}$ to use the inner product of $F(s)$ and $F(t)$ fails. We presume the reason may be that the value of the inner product between two binary vectors can only take integers from $[-L,L]$ which limits the representation power of the model. We therefore concatenate $F(s)$ and $F(t)$ and passing it through a multi-layer perceptron (MLP) to predict the matching label $y$. Interchangeability is still loosely maintained as the pair $(S,T)$ is randomly swapped when feeding into the discriminator.


\paragraph{Utterance pair construction}
One issue in constructing the positive/negative pair for the feature extractor is that the number of positive/negative pairs need to be balanced to achieve a robust empirical result. Moreover, when constructing the positive sample pairs with $y = 1$, if the $s$ and $t$ are adjacent or close to each other in a dialogue, we might end up capturing adjacency pairs~\cite{sacks73}
rather than conversation topics. For example, $s=$  \texttt{'How are you?'} and $t=$ \texttt{'Fine. How are you?'}. The captured similarity in feature space of this $s,t$ pair is contextual appropriateness rather than topic/persona consistency. To alleviate this, we collect only those pairs that are more than 4 turns away from each other for the positive sample pairs.

We note also that the persona features may affect the topic feature extractor because the persona features can be weak signals for predicting whether two sentences are from the same dialogue. One remedy is to select utterances from different speakers within a dialogue session when constructing the positive pairs for the topic extractor to eliminate as much as possible the effect of the persona features. However, this remedy can result in fewer positive pairs. Empirically the topic extractor works well even without this remedy, presumably because the strong signal from topic overwhelms the weak signal from persona.

\begin{figure*}[ht!]
    \centering
    \includegraphics[width=.47\textwidth]{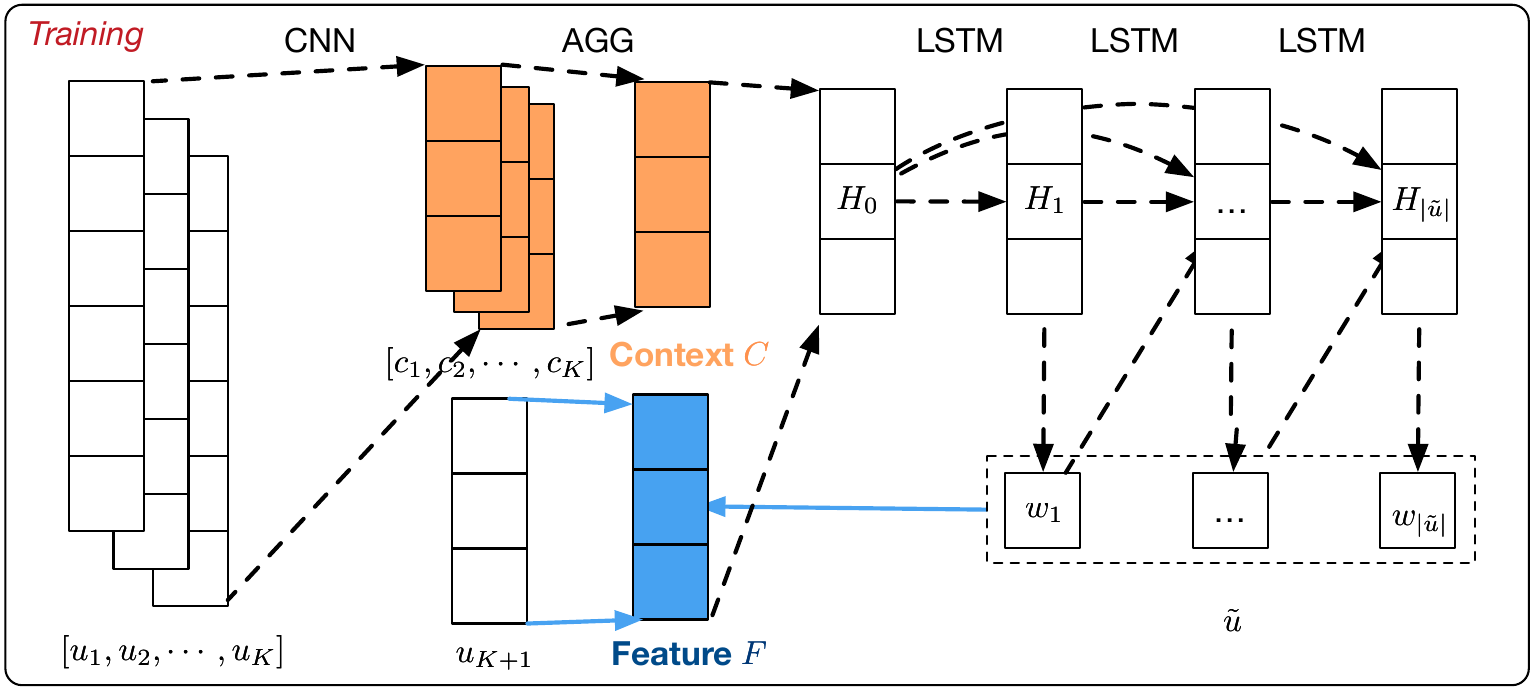}
    \includegraphics[width=.47\textwidth]{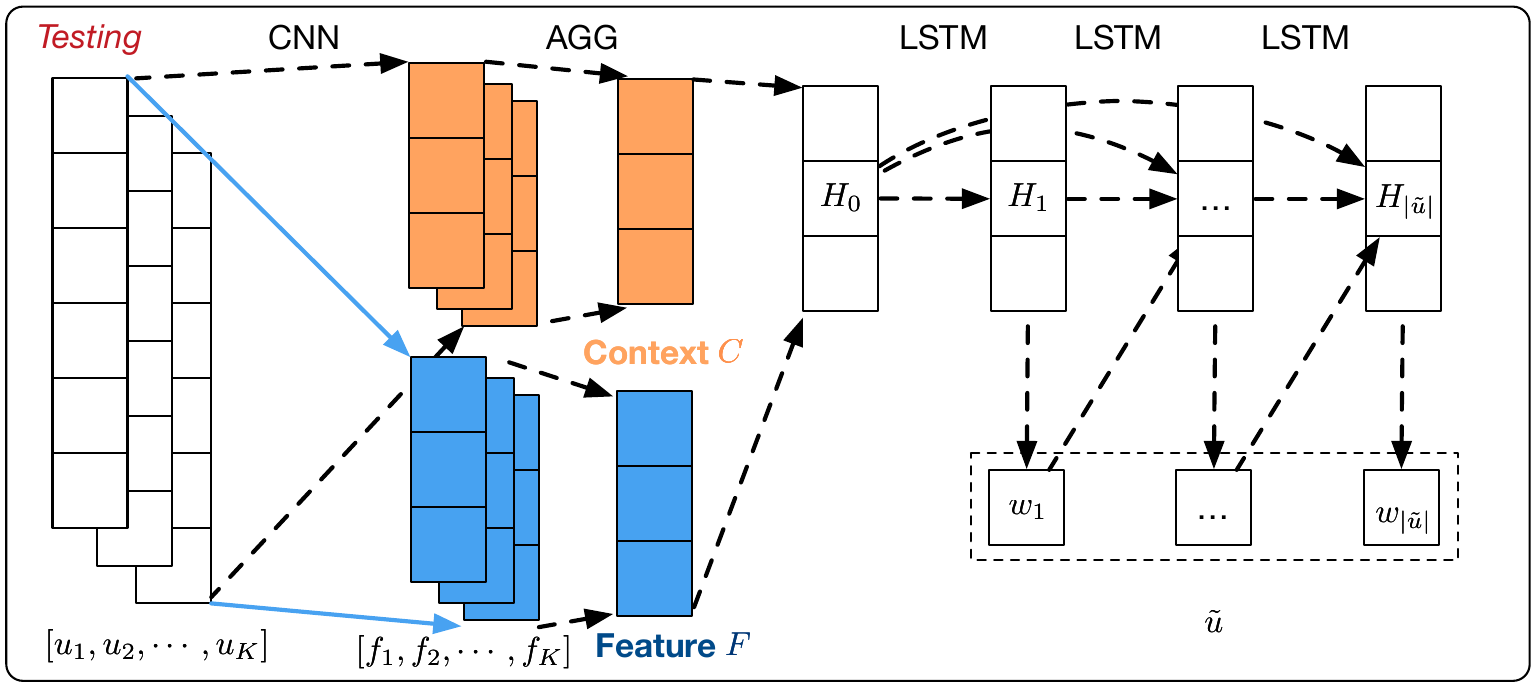}
    \caption{Controllable generation scheme (better in color). The feature extractor $F(\cdot)$ are represented as solid blue arrows. During training time, contextual sources $[u_1,\cdots,u_K]$ are encoded (by encoder), and aggregated (by aggregator) to a context vector $C$, meanwhile the target $u_{K+1}$ is abstracted by feature extractor as feature vector $F$. The decoder then (controllably) generates a response $\tilde{u}$ based on both $C$ and $F$. For testing, the feature $F$ is obtained by aggregating the feature vectors for each source sentence.  }
    \label{fig:g}
\end{figure*}

\subsection{Generator design}

\paragraph{Training a response generator}
The conditional multi-turn generator that produces neural responses given the $K$-turn source sentences is shown in Figure~\ref{fig:g}, which is conceptually related to \cite{serban2015hierarchical}. During \textit{training} time (Figure~\ref{fig:g} left panel), each source sentence is first encoded by a 3-layer CNN encoder, which shares the same architecture as the feature extractor, followed by a context aggregator ($AGG_C$) layer that summarizes all sentence embedding vectors $[c_1,c_2,\cdots,c_K]$ into one single context vector $C$ with the same dimension as $c_i$. In this paper, the $AGG_C$ layer is designed as first concatenating $[c_1,c_2,\cdots,c_K]$  and applying a fully-connected layer to map the resulting vector to $C$.

On the other hand, the target sentence is processed by the feature extractors to produce feature vector(s) $F$ as described in Section~\ref{sec:d}. The feature extractors are fixed in the response generator since we observed fine-tuning the feature extractor leads to suboptimal empirical results. The context vector $C$ and feature vector(s) $F$ are fed into an MLP to generate a fixed-length initial hidden variable $H_0$. This is followed by a series of long short-term memory (LSTM) units as the decoder, where $H_0$ is employed as input in each time-step. 

\paragraph{Controllable objective during training} 
Our generator loss incorporates two components. The first one is the vanilla teacher-forcing \cite{williams1989learning} MLE loss $\mathcal{L}_\texttt{MLE}$. The second part is a cycle consistency loss $\mathcal{L}_\texttt{cycl}$, introduced by \cite{hu2017toward} to admit additional controlling ability of feature vector in the generation. Intuitively, it encourages the self-generated response under free-running generative mode to have the same features as the input signal $F$. Specifically, consider a response $\tilde{u}=[w_1, w_2 ,\cdots, w_{|\tilde{u}|}]$ greedily generated by conditioning on previously generated tokens. The $\mathcal{L}_\texttt{cycl}$ is simply the Euclidean distance between input feature vectors $F$ and $F(\tilde{u})$, \ie $\mathcal{L}_\texttt{cycl}=||F-F(\tilde{u})||^2$. In the case of binary features, $\mathcal{L}_\texttt{cycl}=||F-P(\tilde{u})||^2$ where $P(\cdot)$ is the network output before rounding to binary values. 
Note that the generated tokens $[w_1, w_2 ,\cdots, w_{|\tilde{u}|}]$ involves an \texttt{argmax} operation and are not directly differentiable, preventing the gradient signals from back-propagating to the encoder and decoder. Common remedies for this includes Gumbel-softmax (GS)\cite{gumbel1954statistical}, policy gradient (PG)\cite{yu2016sequence} and soft-argmax (SA)\cite{zhang2017adversarial}. Unfortunately, GS and PG suffer from high variances of gradient estimation while SA suffers from a dilemma between gradient vanishing and inaccurate gradient.
To alleviate such a problem in SA, we consider an approach in what we call the {\it Straight-Through} LSTM unit (ST-LSTM), which use ST estimation~\cite{bengio2013estimating, jang2016categorical} to achieve a biased but smooth gradient signal while maintaining the forward computation exact via a temperature parameter $\tau$. The details are provided in the Appendix. 


In the experiment, we applied the slope-annealing trick \cite{chung2016hierarchical}, and set $\tau=0.01$ which works well in practice. The final training objective for the generation is $\Lcal_\texttt{MLE} + \eta \mathcal{L}_\texttt{cycl}$.

\paragraph{Testing time} 
At test time, as shown in Figure~\ref{fig:g} (right panel), the feature vectors from the source sentences $[u_1,\cdots,u_K]$ are first collected by applying feature extractors $F(\cdot)$. We denote the feature vectors for the source sentences as $[f_1, \cdot, f_K]$. We apply a feature aggregator $AGG_F$ layer to estimate the output feature vector $F$, which is further fed into the LSTM-RNN for the generation. Different from the context $AGG_C$ layer, we consider a weighted-sum aggregation function for the feature $AGG_F$ layer \footnote{Other possibilities of such an $AGG_F$ layer exist, such as mean, max or concatenation (as in $AGG_C$). We choose weighted-sum for the $AGG_F$ layer due to its superior empirical performance comparing to alternatives.}, \ie, $F = \sum_{k=1}^K w_k f_k, s.t. \sum_{k=1}^K w_k = 1$, where $w_k, k=[1,2,\cdots,K]$ are linear interpolation weights learned during training time, where a Euclidean distance between predicted target feature and target feature is optimized, \ie $\text{argmin}_{w} \Lcal_p = ||f_{k+1}-\sum_{k=1}^K w_k f_k||^2$. 
For the persona feature, we only use the source sentences of the current speaker, thus all $w_k$ where $\texttt{mod}(k,2)=\texttt{mod}(K,2)$ is set as zero. Intuitively it can be perceived as the attention of each utterance.
We note that more complicated attention mechanisms can further improve the model; however, we leave these for future work, since this paper focuses on the utilization of dialogue features rather than improving the multi-turn S2S structure in general. 



\section{Experimental setups}
We evaluate the proposed methods on two datasets. All experiments are conducted using single Nvidia Tesla V100 GPU. The source code will be released. 
\subsection{Data collection}
We consider two datasets. For both we use a (80\%, 10\%, 10\%) split for training, validation and test respectively. 
\label{sec:res}
\paragraph{Twitter data} 
Training data was extracted from the Twitter FireHose covering a five-year period from 2012 through 2016.\footnote{Deleted tweets and closed accounts were removed.} From this set, we collected total 6,658,385 8-turn dialogues where two participants chatted with each other. 

\paragraph{Maluuba data} 
The Maluuba dataset consists of 40,389 dialogues with 11 turns. Each dialogue is a task-oriented conversational interaction between two real speakers regarding 51 domains and 242 tasks, collected by crowd-sourcing where one crowd worker simulates a user and another simulates a chatbot. 

\subsection{System specifications}
The dimension of the LSTM hidden layer is set at 500. We use ADAM as the optimizer with learning rate 0.0001. The hyperparameters $\lambda$ and $\eta$ are set at 0.01 and 0.1, respectively. For the dimension of feature vectors we use 100.  For Maluuba dataset we use a 50\% dropout rate in each of the CNN layers and the $\lambda$ is set to $0.1$. The hyperparameters are selected to maintain the discrimination accuracy while reducing as much $\Lcal_\texttt{DeCorr}$ as possible. 

For evaluation, we consider three variants of our COnsistent CONversation (CoCon) models: \textbf{CoCon-T}: CoCon model with topic-consistency; \textbf{CoCon-TP}: CoCon model with topic-consistency and persona-consistency; \textbf{CoCon-TP-bin}: CoCon model using binary features with topic-consistency and persona-consistency. 
We compared our models with two baselines: a vanilla sequence-to-sequence model (\textbf{S2S}) and persona model (\textbf{Persona}) \cite{li2016persona}. We implement the persona model by reusing the encoder and decoder architecture in our approach. For Twitter dataset, we map all users with fewer than 88 utterances as unknown (86\% of the total training samples) and in the test set (for all compared methods) we eliminate conversation sessions with unknown users. This yields 50k total users. 
We use the same number of feature dimensions for all systems compared. All modules are trained until convergence.

\section{Results}

\paragraph{Self-supervised feature learning}
We used equal numbers of positive/negative examples to train each feature extractor. For Twitter dataset, the resulting accuracies for topic and persona feature extractor are around $0.75$ and $0.60$ (for both continuous and binary features), respectively. For Maluuba dataset, the discriminator accuracy for persona and topic feature extractors are $0.85$ and $0.67$, respectively. With the disentangling loss ($\lambda=0.01$), the correlation between features drops from 0.25 to 0.16.

Representative n-grams for some learned feature units for Twitter dataset are shown in Table~\ref{tab:topic}. To calculate the feature vector for a specific n-gram, we average over the feature vector of test sentences that contain that n-gram. We then select the top-ranked n-grams with occurrences greater than 200 for each feature bit. 
We observe that when $\lambda=0$, \ie without disentangling loss, the learned features exhibit heavy colinearity, which weakens the interpretability of each separate feature units.



\begin{table}[ht!]
\vspace{4mm}\scriptsize
\begin{tabular}{p{.3in}|p{.3in}p{.45in}p{.5in}p{.7in}}
Topics & 1gram & 2gram  & 3gram  & 4gram\\
\hline
T-2 (electronic) & android; ios; apps & the iphone; the app; my ipad & a new phone; my phone is; are you using & is on my phone; you can use the; send it to the\\
\hline
T-59 (sport) & striker; arsenal; madrid & champions league ; best player; the spurs; & in the league; in the playoffs; a good game & one of the best; best of the season; the team in the; \\
\hline
T-18 (Movie , show) & episodes; film; netflix & of thrones; that movie; the ending & i watched it; to watch it; the first one & one of my favorite; have you seen the; i want to see\\
\hline
\end{tabular}

\scriptsize
\begin{tabular}{p{.3in}|p{.3in}p{.4in}p{.5in}p{.7in}}
\hline
Persona & 1gram & 2gram  & 3gram  & 4gram\\
\hline
P-71 (inquiry) & what; thx; wheres & can u; do u; where is & how do i; is there a; what is the & thank you so much; i need to get ; i want to see\\
\hline
P-49 (agreement) & yea; great; sure & lol yea; yea i; hell yea & that s so; i will do; yea i m & i would love to; looking forward to it; sounds like a plan \\
\hline
P-28 (abbreviation) & hav; urself; wats &u will; u got; for u & i thought u; did u get; do you have & s what s up; i m bout to; do u have any\\
\hline
P-83 (african amer.) & aint; homie; yo; &n**** i;hell yea;i ain & u ain 't; cuz i am; i 'm tryna & i 'm bout to; i 'm gonna have; ain 't nothing gonna \\
\end{tabular}
\caption{Representative n-grams in topic (T) and persona (P) features}
\label{tab:topic}
\vspace{-2mm} 
\end{table}

\begin{table}[ht!]
\scriptsize
\begin{tabular}{p{0.5in}|p{2.2in}}
\hline
\textbf{Context}:& {A:tommorows game just became that much bigger . must win .}\\
& {B: unbeliebable . mourinho has never lost at home in the \textbf{league} as chelsea \textbf{coach} and against the worst \textbf{team} in the \textbf{league} he loses}\\
& {A:yup , would be insane if they could somehow avoid relegation now .  }\\
& {B:would be amazing.we played against them at home 3 weeks ago but barely beat them . they are always up for it against the big \textbf{teams}}\\
\hline
\textbf{Human}:& {A:yeah , said they played aggainst top half of table much better then bottom half .}\\

\hline
\textbf{S2S}:& {A:i think they 'll be fine . . .}\\
\hline
\textbf{Persona}:& {A:i think they 'll be in the same position .   }\\
\hline
(Ours) \textbf{CoCon-T}:& {A:i think they \textbf{play} well . \textbf{i 'm a big fan of the manager} . they 've \textbf{played} well in the same \textbf{league} as well .}\\
\hline
(Ours) \textbf{CoCon-TP}:& {A:yeah , they \textbf{played} well with the top three \textbf{teams} , but i 'd say it was . not sure what to expect . }\\
\hline
\end{tabular}

\begin{tabular}{p{0.5in}|p{2.2in}}
\hline
\textbf{Context}:& {A:lucky you \textbf{mate} . have a super fine \textbf{weekend} buddy}\\
& {B: i m gonna be knackered next week should be a good \textbf{weekend} though gonna get loads a pics}\\
& {A:can n't fault you \textbf{mate} nice mid night start tuesday lol }\\
& {B:gonna be at the show saturday then home sunday back to work monday . where you off to tomorrow ?}\\
\hline
\textbf{Human}:& {A:one out of cedar farm to march then not sure mate either ipswich or a hiams to march .}\\
\hline
\textbf{S2S}:& {A:not sure . i 'm off for a few days}\\
\hline
\textbf{Persona}:& {A:i 'm off friday night . you going to the game ?  }\\
\hline
(Ours) \textbf{CoCon-T}:& {A:i know . i am going to manchester on the \textbf{weekend} . what about you ?  }\\
\hline
(Ours) \textbf{CoCon-TP}:& {A:i 'm going to the beach \textbf{mate} . i 'm going for a couple days off . \textbf{good weekend pal} }\\
\hline
\end{tabular}
\caption{Sampled generation on Twitter. }\label{tab:examples}
\end{table}

 \begin{figure}[ht!]
     \centering
     \includegraphics[width=.48\textwidth]{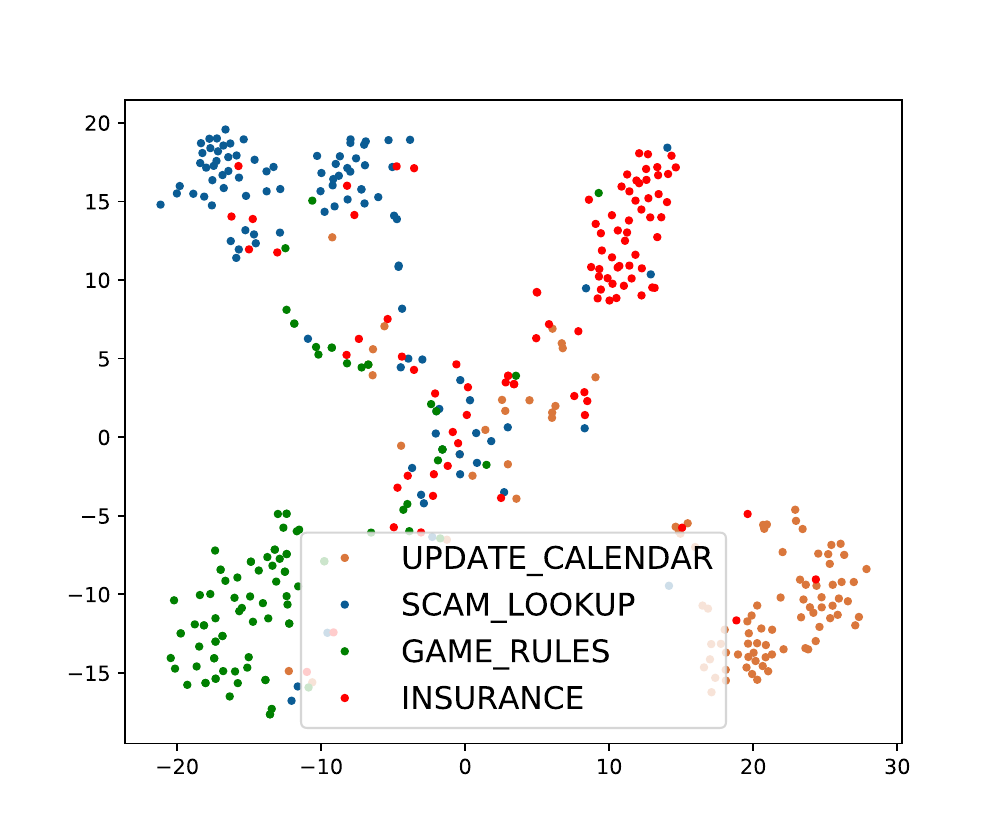}
      \includegraphics[width=.48\textwidth]{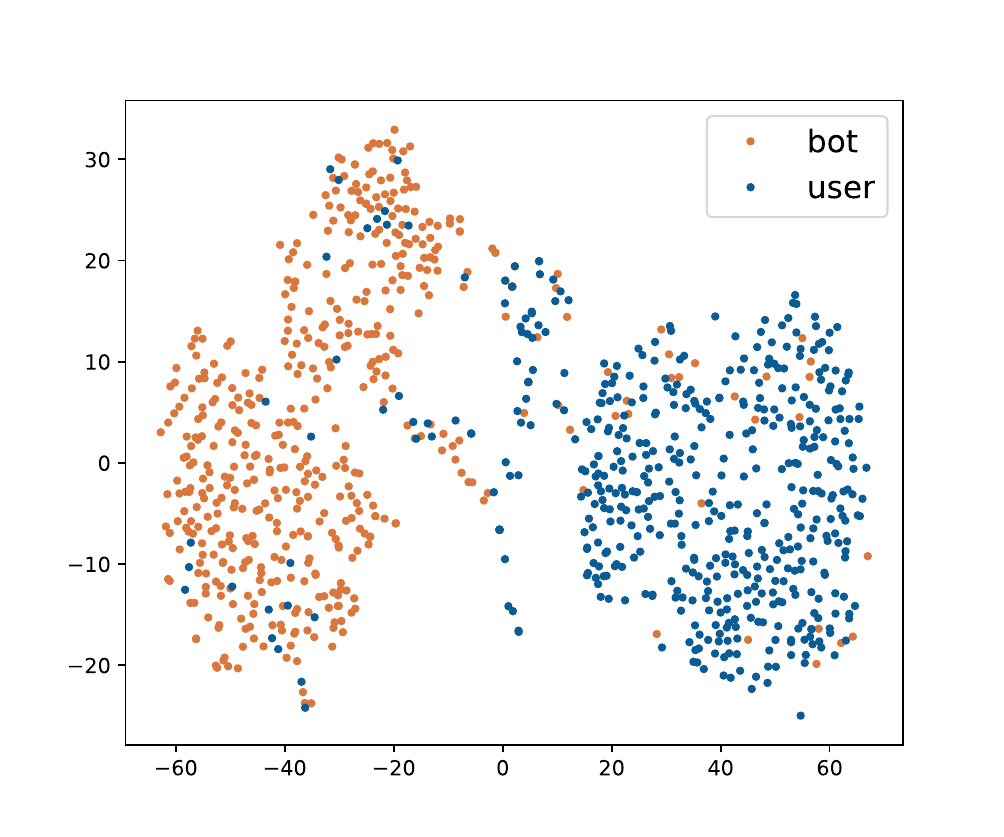}
     \caption{t-SNE embedding of topic (left) and persona (right) features for different tasks (left) and roles (right).}
     \label{fig:p_emb}
 \end{figure}



\begin{table}[ht!]
\scriptsize
\begin{tabular}{p{0.3in}|p{2.5in}}
\hline
\textbf{Context}:& {A:done with the \textbf{game}}\\
& {B:what are you talking about}\\
& {A:you heard me}\\
& {B:if you 're not there today i 'm going to be mad}\\
\hline
\textbf{Human}:& {A:lol ok i aint going}\\
& {B:are you trying to be a bod mon or something ?}\\
& {A:oh you that 's exactly how i 'm tryna be aha}\\
& {B:so tell me why you are n't going ? you just say you 're not , talk about it}\\
\hline
\textbf{Plain}:& {A:i'm on the vacation. i'm not going to \textbf{play that game}}\\
& {B:you don't really mean this, right? you're n't going to do that}\\
& {A:i aint , i told you . }\\
& {B:well you can n't be a person like that , you can n't be a loser}\\
\hline
\textbf{Sport}:& {A:i'm watching the school \textbf{football}. that's enough for now}\\
\texttt{on}& {B:that's why you're a loser }\\
& {A:lol no i'm going to be the best \textbf{player} in the \textbf{league}.}\\
& {B:oh well you're not going to win man you are terrible at all this time.}\\
\hline
\end{tabular}
\caption{Examples of bit manipulations. \textit{Plain} denotes without manipulation. $X$ \texttt{on} denotes bit $X$ is activated (set to $1$) when generating the response }
\label{tab:ctrl}
\end{table}



We further visualized the topic features on both datasets using t-SNE embedding~\cite{maaten2008visualizing}. For Maluuba dataset, Figure~\ref{fig:p_emb} illustrates the learned topic and persona feature embeddings on the test set. Without any label information, the learned topic and persona features separate well. For twitter dataset, we observed that the persona features of the utterances from different time zones form some clusters, indicating the features learned from our approach can partially reflect the difference in societal groups (See Appendix~\ref{app:twt}). 



\paragraph{Sampled response generation}
We evaluate our approaches by generating the next response given 4 contextual seed source sentences. Some sampled results are shown in Table~\ref{tab:examples}. We observed that the CoCon-T and CoCon-TP in general are able to produce informative responses which seem to be more consistent with the theme of the given context comparing with baselines. For CoCon-TP, beyond being context-aware, the responses seem to be persona-aware, \ie,  mimicking the tone and personal wording preferences like \texttt{mate}, \texttt{oh my gosh}, \texttt{haha}, \texttt{ain 't} and other words associated with them.

\paragraph{Feature manipulation}
We further manipulate feature bits that seem to be associated with certain topics. The results are shown in Table~\ref{tab:ctrl} (additional results are provided in the Appendix). We generate next 4 turns consecutively. The later generations consider the previous 4 sentences, including previously generated utterances as source context. This bit manipulation is based on binary feature codes, achieved by toggling the specific bit to be $1$ to activate it. With the additional controllable generation objective $\Lcal_{\texttt{cycl}}$, we are able to better control the flipping of each bit. As shown in Figure~\ref{fig:z_loss} in Appendix~\ref{app:twt}, increasing $\eta$ leads to a fast decrease of $\Lcal_{\texttt{cycl}}$ (indicating a better controlling power), however may at a cost of harming the generation quality. We select the $\eta=0.01$ by trading off between both aspects. We observe the success rate of bit toggling is about $17\%$ percent (based on 2000 tested cases), meaning that around $17\%$ cases where we flip a $0$ of the input feature $F$ to $1$, the response feature $\tilde{F}$ will remain $1$. Presumably, the model has learned to detect that, based on the context, it is \textit{unnatural} to toggle a certain bit and refused to make the change. This hypothesis needs further experimental verification. Controllable generation is still an emergent technology and the noisy nature of dialogue data makes this even more challenging. 

For Maluuba dataset, we provide sampled responses of S2S and CoCon-TP in Table~\ref{tab:sample_maluuba} in Appendix~\ref{tab:evaluation_maluuba}. The context is given as 4 turns of dialogues and the task to generate all remaining 7 turns. It can be seen that during free generation, the S2S model tend to generate looping responses like \texttt{thanks} -  \texttt{you 're welcome} and is generally less informative. However, our proposed CoCon-TP approach can generate reasonably well by unsupervisedly capturing the topics of the context and role of each turn. 

\begin{table*}[ht!]
  \centering
  \small
  \begin{tabular}{c|c|c|c|c|c|c|c|c|c|c}
    \hline
    \multicolumn{2}{c|}{\multirow{2}{*}{Models}} & \multicolumn{6}{c|}{Relevance} & \multicolumn{3}{c}{Diversity}\\
    \hhline{~~---------}
    \multicolumn{2}{c|}{}	&BLEU & METEOR & NIST & Greedy	&Average	&Extreme 	&Dist-1	&Dist-2	&Ent-4 \\
    \hline
    \multirow{2}{*}{Ours} &CoCon-T 	&3.01	& 0.061	&  1.022	& 1.968  & 0.675	& 0.321 & 0.008 & 0.065	& 9.71\\
    &CoCon-TP 	&\textbf{3.31}	& \textbf{0.064}	&  \textbf{1.135} &\textbf{ 2.048}  & \textbf{0.683}	& \textbf{0.342 } & 0.008	& 0.081	& 10.46\\
    &CoCon-TP-bin 	&3.04	& 0.063	& 1.061 & 2.025  & 0.677	&  0.331 &\textbf{ 0.009} &\textbf{0.100} & \textbf{10.59}\\
    \hline
    \multicolumn{2}{c|}{S2S}   & 2.83 & 0.056 & 0.945	& 1.855	& 0.640	&0.307 	& 0.004	& 0.023	& 7.51 \\
    \multicolumn{2}{c|}{Persona model} 	& 2.96 	& 0.059 	& 1.014	&  1.931 	& 0.658 &  0.319 	& 0.005	& 0.028 & 7.96 \\
    \hline 
    \multicolumn{2}{c|}{Human} 	& - 	& - 	& -	& -  &-	& - & 0.078 &0.473	& 11.75\\
    \hline
  \end{tabular}
  \caption{Quantitative evaluation for twitter dataset}\label{tab:evaluation_twt}
\end{table*}
	
\begin{table*}[ht!]
  \caption{Quantitative evaluation for Maluuba dataset}\label{tab:evaluation_maluuba}
  \centering
  \small
  \begin{tabular}{c|c|c|c|c|c|c|c|c|c|c}
    \hline
    \multicolumn{2}{c|}{\multirow{2}{*}{Models}} & \multicolumn{6}{c|}{Relevance} & \multicolumn{3}{c}{Diversity}\\
    \hhline{~~---------}
    \multicolumn{2}{c|}{}	&BLEU & METEOR & NIST & Greedy	&Average	&Extreme 	&Dist-1	&Dist-2	&Ent-4 \\
    \hline
    \multirow{2}{*}{Ours} &CoCon-T 	&5.6  & 0.076 &	 1.421	& 2.105	&   0.565	&  0.358  & 0.025	& 0.142	&  8.899\\ 
    &CoCon-TP 	&\textbf{5.8}  & \textbf{0.077} & \textbf{1.459} & \textbf{2.175}	&  \textbf{0.575}	&\textbf{0.365}  & \textbf{0.028} 	& 0.16	& 8.983\\      
    &CoCon-TP-bin 	&4.6  & 0.074 &	1.280	& 2.094	&  0.559	& 0.341  & 0.027 	& \textbf{0.19}	& \textbf{9.767}\\
    \hline
     \multicolumn{2}{c|}{S2S}  & 3.9  & 0.066 & 1.045	& 2.021 	& 0.529	& 0.328 	& 0.017 & 0.16 	&8.293\\
    \multicolumn{2}{c|}{Persona model} 	& 4.4 	& 0.073	& 1.134 & 2.042	& 0.543	& 0.319 & 0.021	& 0.177	& 8.603\\
    \hline 
    \multicolumn{2}{c|}{Human} 	& - 	& - 	& - 	& -	& - 	& - &0.092 	& 0.462	& 10.281\\
    \hline
  \end{tabular}
\end{table*}

\begin{table*}[ht!]
\small
\centering
\begin{tabular}{r r | r| r l  l r r | r | r | l}
\cmidrule[\heavyrulewidth]{1-11}
 \multicolumn{5}{c}{Topic Consistency (human judges preferred) } & & \multicolumn{5}{c}{Persona Consistency (human judges preferred)}\\
\cmidrule[\heavyrulewidth]{1-5} \cmidrule[\heavyrulewidth]{7-11}
\multicolumn{2}{c|}{Our Method} & Neutral & \multicolumn{2}{c}{Comparison} & & \multicolumn{2}{c|}{Our Method} & Neutral & \multicolumn{2}{c}{Comparison} \\ 
\cmidrule[\heavyrulewidth]{1-5} \cmidrule[\heavyrulewidth]{7-11}
\cmidrule{1-5} \cmidrule{7-11}
CoCon-TP & \bf{45.20}\% & 22.30\% & 32.50\% & seq2seq && CoCon-TP & \bf{40.95}\% & 29.85\% & 29.20\% & seq2seq\\
CoCon-TP & \bf{40.05}\% & 23.10\% & 36.85\% & persona && CoCon-TP & \bf{35.65}\% & 34.10\% & 30.25\% & persona \\
\cmidrule{1-5} \cmidrule{7-11}
CoCon-TP & 21.50\% & 26.85\% & \bf{51.65}\% & human && CoCon-TP & 21.35\% & 33.35\% & \bf{45.30}\% & human \\
	\cmidrule[\heavyrulewidth]{1-11}
	\end{tabular}
\caption{Results of {\bf Human Evaluation} for topical and persona consistency, showing preferences (\%) for our model (CoCon-TP) vis-a-vis baseline or other comparison systems. Distributions are skewed towards CoCon-TP, except when compared with human outputs. Numbers in bold indicate the most preferred systems. For simplicity, the 5-point Likert scale is collapsed to a 3-point scale. 
See the Appendix for further details. }\label{tab:human_eval}

\end{table*}

\paragraph{Automatic evaluations}
In our quantitative evaluations we test both \textit{relevance} and \textit{diversity} metrics. For relevance, we adopt \textbf{BLEU}~\cite{papineni2002bleu}, \textbf{METEOR}~\cite{denkowski2014meteor}, \textbf{NIST}~\cite{doddington2002nist} and three embedding-based metrics \textbf{Greedy},\textbf{Average},\textbf{Extreme} following \cite{serban2017hierarchical,rus2012comparison,mitchell2008vector,forgues2014bootstrapping}. 
To evaluate diversity, we follow \cite{li2015diversity} to use \textbf{Dist-1} and \textbf{Dist-2}, which is characterized by the proportion between the number of unique n-grams and total number of n-grams of tested sentence. We also include the \textbf{Entropy} (Ent-n) metric \cite{zhang2018generating, gao2019jointly}, which does not depend on the size of test data.
The results of automatic evaluations are shown in Table~\ref{tab:evaluation_twt} (Twitter) and Table~\ref{tab:evaluation_maluuba} (Maluuba). For both dataset, the CoCon-TP model achieves best relevance score, while the CoCon-TP-bin outperforms other methods in diversity.

 \paragraph{Human evaluations}

We evaluated 500 randomly sampled test sources from Twitter dataset using crowd-sourcing provided by a contracting service. Systems were paired and each pair of system outputs was randomly presented to 4 judges, who ranked them for topic consistency, persona consistency, informativeness and relevance using a 5-point Likert scale. Overall judges' preferences for the topic consistency, persona consistency, given as a percentage of total judgments are shown in Table \ref{tab:human_eval}. A strong overall preference can be observed for CoCon-TP over the other systems evaluated. We also evaluated for relevance and informativeness, with CoCon-TP showing similar preference gains. Further details, including the human evaluation template used, are provided in the Appendix. 





\section{Conclusion}
We present a self-supervised feature learning framework to abstract high-level latent representations of topic and persona information underlying the dialogue context and leverage these representations to generate more consistent dialogue in a controllable manner. For future work, investigating the variance reduction strategies for controllable text generation would presumably improve the controllablity of the feature units. Besides, combining and aligning supervised and unsupervised features would potentially enable better feature learning and interpretability.
Our approach can be adapted to facilitate style transfer and long-form text generation \cite{guo2017long, zhang2017adversarial} to improve the generation consistency.  
\bibliography{main}
\bibliographystyle{acl_natbib}

\clearpage
\newpage
\appendix

\twocolumn[\centering{\Large \bf Appendix for Consistent Dialogue Generation with Self-supervised Feature Learning}\vspace{5mm}]

\vspace{5mm}

\section{Straight-through LSTM (ST-LSTM)}
In the forward calculation, the $t$-th ST-LSTM unit takes the previously generated word $w_{t-1}$, hidden state $H_{t-1}$ and $H_0$ as input, and generates the next word $w_t$ that maximizes the probability $p_t$ over the vocabulary set. That is, the {\it argmax} operation is used. However, in the backward calculation, the gradient of $\partial w_t / \partial p_t$ is calculated as a constant $1/\tau$ where $\tau$ is a temperature parameter.
Presumably, this approach delivers a biased but smooth gradient signal while maintaining the forward computation exact.

\section{Twitter additional results}
\label{app:twt}
\begin{figure}[ht!]
    \centering
    \includegraphics[width=.9\textwidth]{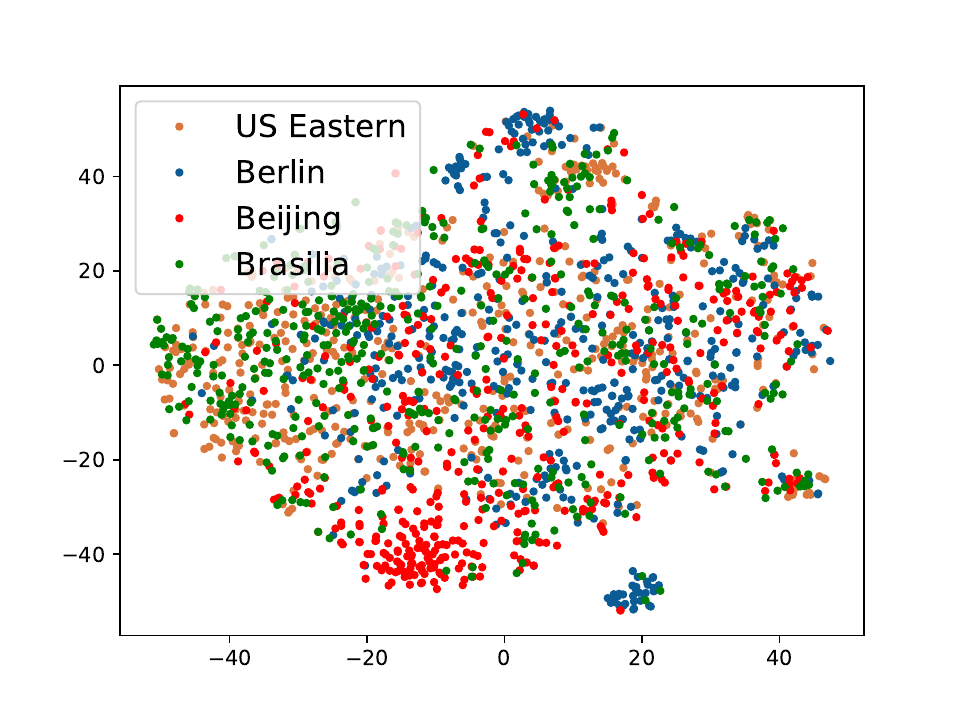}
    \caption{t-SNE embedding of topic features for speakers in different time zones.}
    \label{fig:twt_topic}
\end{figure}

Figure~\ref{fig:twt_topic} shows the t-SNE embedding of topic features for speakers in different time zones, demonstrating our self-supervised approach learned reasonable representation to separate different users. Here we use 2000 testing utterances with speakers from 4 different time zones (500 for each). 

\begin{figure}[ht!]
    \centering
    \includegraphics[width=.9\textwidth]{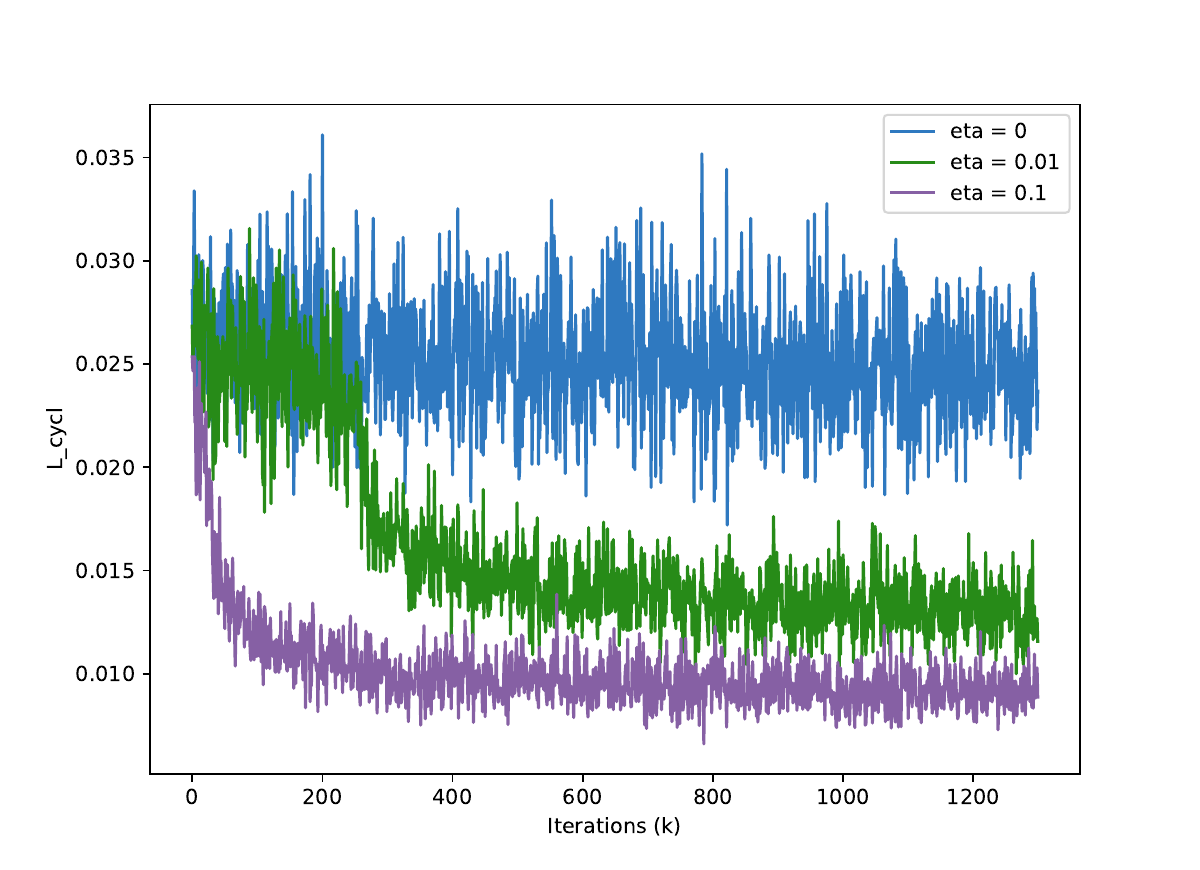}
    \caption{$\Lcal_{\texttt{cycl}}$ decreases faster when $\eta$ is larger.}
    \label{fig:z_loss}
\end{figure}

Figure~\ref{fig:z_loss} shows that $\Lcal_{\texttt{cycl}}$ decreases faster when $\eta$ is larger, indicating the additional controlling loss  $\Lcal_{\texttt{cycl}}$ can be effectively reduced using Straight-through training. 

Table~\ref{tab:ctrl2} shows some additional results on controlling the bit, in addition to Table~\ref{tab:ctrl}. 
\begin{table}[ht!]
\scriptsize
\begin{tabular}{p{0.3in}|p{2.5in}}
\hline
\textbf{Context}:& {A:planning on making a trip to hawaii}\\
& {B:lol if only people knew what we are really going there for lolol}\\
& {A:hahaha . you we might get judge a little}\\
& {B:a little ? probably a lot haha}\\
\hline
\textbf{plain}:& {A:it's okay. we should totally go together. we can become friends}\\
& {B:lol i'm not sure. i'm going to vegas instead!}\\
& {A:haha. i feel you. i just want to get out of all of these.}\\
& {B:let's get on our way!!}\\
\hline
\textbf{electronic \texttt{on}}:& {A:it's okay. we should totally go together. i will friend you on \textbf{fb}}\\
& {B:lol sounds good. i will \textbf{use my phone}}\\
& {A:sure. let me add you to my \textbf{contact}}\\
& {B:i'm going to do that. i will add you too.}\\
\hline

\end{tabular}

\caption{Examples of bits manipulations. \textit{Plain} denotes without manipulation. $X$ {on} denotes bit $X$ is activated (set to $1$) when generating the response }
\label{tab:ctrl2}
\end{table}

\begin{table}[ht!]
\scriptsize
\begin{tabular}{l|p{2.3in}}
\hline
\textbf{Source}:& {A:hello how may i help you ?}\\
& {U:hi i want to look up a play}\\
& {A:what play}\\
& {U:what time does the secret miracle opening on ?}\\
\hline
\textbf{Human}:& {A:the secret miracle is scheduled for 8 pm on saturday}\\
& {U:ok book me for tickets}\\
& {A:ok sure}\\
& {U:how much is that ?}\\
& {A:50}\\
& {U:ok book it now}\\
& {A:done}\\
\hline
\textbf{CoCon-TP}:& {A:9:00 PM friday}\\
& {U:are there any tickets available for saturday there ?}\\
& {A:yes tickets do have a deal at 8 pm}\\
& {U:ok book me for tickets}\\
& {A:ok ticket}\\
& {U:reserve for 3 seats please}\\
& {A:okay done}\\
\hline
\textbf{S2S}:& {A:it 's on saturday}\\
& {U:and what time ?}\\
& {A:2:00 PM}\\
& {U:\textbf{thank you}}\\
& {A:\textbf{you 're welcome}}\\
& {U:\textbf{thanks for the help}}\\
& {A:\textbf{you 're welcome}}\\ 
\hline
\end{tabular}

\begin{tabular}{l|p{2.3in}}
\hline
\textbf{Source}:& {A:hello how may i help you ?}\\
& {U:i need to learn more about the rules for tag}\\
& {A:i can certainly help with game rules . what do you want to learn about the game tag ?}\\
& {U:how do i pick the first person to be ' it '}\\
\hline
\textbf{Human}:& {A: the first person to ask who is it is considered ' it '. }\\
& {U:hm , that makes sense . how do you know who is the winner ?}\\
& {A:you could keep track of times a person has been ' it '}\\
& {U:oh , okay thanks so much}\\
& {A:can i help you with something else today ? }\\
& {U:no , that will be all today}\\
& {A:thanks for taking the time to ask me a question !}\\
\hline
\textbf{CoCon-TP}:& {A:the game of tag is a person who gets tagged the least .}\\
& {U:is there a different rule for tag ?}\\
& {A:yes , i can help you with that . there is a lot of tag}\\
& {U:okay , what is the best time to play the most ?}\\
& {A:the winner is the person that gets tagged the least }\\
& {U:oh , okay , i 'll play that game . thanks for the tips .}\\
& {A:you are welcome . i am always here when you need to learn .}\\
\hline
\textbf{S2S}:& {A:the person who asks who is it is declared ' it '.}\\
& {U:\textbf{okay , thank you}}\\
& {A:\textbf{you 're welcome }.}\\
& {U:\textbf{okay , thanks}}\\
& {A:\textbf{you 're welcome !}}\\
& {U:have a good day}\\
& {A:you too .}\\ 
\hline
\end{tabular}

\caption{Generated Responses for maluuba dataset (in consecutive generation manner)}
\label{tab:sample_maluuba}
\end{table}

\begin{figure*}[ht!]
    \centering
    \includegraphics[width=0.8\paperwidth,keepaspectratio]{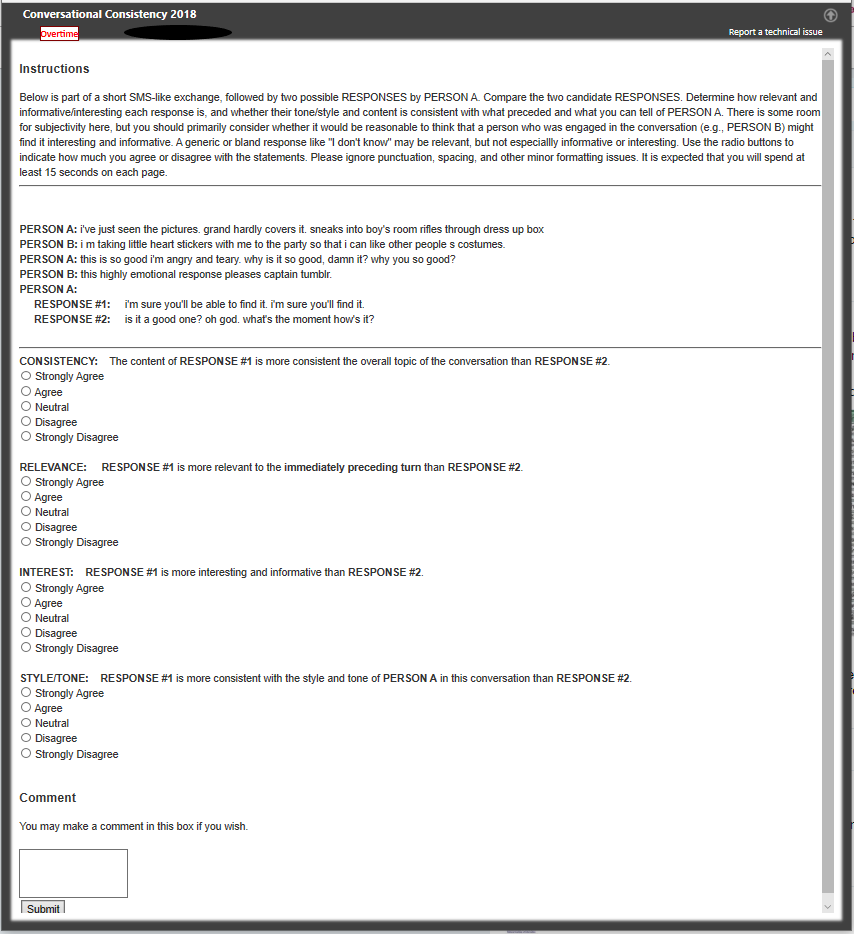}
    \caption{Questionnaire used to elicit pairwise judgments from crowdsourced annotators. Candidate responses were presented in random order.}%
    \label{fig:questionnaire}%
\end{figure*}

\section{Maluuba results}
We provide some generated samples for maluuba dataset in Table~\ref{tab:sample_maluuba}. All compared models use first 4 turns as seed and generate the remaining 7 turns by taking 4 previous \footnote{previous turn use ground truth for first 4 and generated utterances for the rest} turns as context. 
\label{sec:maluuba}


\section{Human evaluation}

Human evaluation was conducted using the form shown in Figure~\ref{fig:questionnaire}. The two response candidates were presented in random order to the judges, who used a Likert scale to indicate their preferences. To make the questionnaire less abstract to judges, persona was evaluated in terms of which response better reflected the tone and style of Person A as observable in the prior turns. The distributions of judgments for each of the questions are shown in Tables~\ref{tab:topic_consistency} through \ref{tab:informativeness}.

\begin{table*}[t!]
\small
\centering
\begin{tabular}{r r | r| r | r | r | l}
\cmidrule[\heavyrulewidth]{1-7}
 \multicolumn{7}{c}{Distribution of Pairwise Topic Consistency Preferences}\\
\cmidrule[\heavyrulewidth]{1-7} 
Our Method & \multicolumn{1}{c|}{5} & \multicolumn{1}{c|}{4} & \multicolumn{1}{c|}{3} & \multicolumn{1}{c|}{2} & \multicolumn{1}{c|}{1} & Baseline \\ 
\cmidrule[\heavyrulewidth]{1-7} 
CoCon-TP & 12.60\% & 32.60\% & 22.30\% & 24.45\% & 8.05\% & seq2seq\\
CoCon-TP & 10.20\% & 29.85\% & 23.10\% & 28.75\% & 8.10\% & persona\\
\cmidrule{1-7} 
CoCon-TP & 5.10\% & 16.40\% & 26.85\% & 33.70\% & 17.95\% & human \\
	\cmidrule[\heavyrulewidth]{1-7}
	\end{tabular}
\caption{Distribution of topical consistency preferences (\%) for our model (CoCon-TP) compared with seq2seq and persona baselines, according to a five-point Likert scale. A 5 indicates a strong preference for CoCon-TP; a 1 indicates strong preference for the alternative system.  }\label{tab:topic_consistency}
\end{table*}

\begin{table*}[t!]
\small
\centering
\begin{tabular}{r r | r| r | r | r | l}
\cmidrule[\heavyrulewidth]{1-7}
 \multicolumn{7}{c}{Distribution of Pairwise Persona Preferences}\\
\cmidrule[\heavyrulewidth]{1-7} 
Our Method & \multicolumn{1}{c|}{5} & \multicolumn{1}{c|}{4} & \multicolumn{1}{c|}{3} & \multicolumn{1}{c|}{2} & \multicolumn{1}{c|}{1} & Baseline \\ 
\cmidrule[\heavyrulewidth]{1-7} 
CoCon-TP & 11.30\% & 29.65\% & 29.85\% & 22.50\% & 6.70\% & seq2seq\\
CoCon-TP & 8.30\% & 27.35\% & 34.10\% & 23.70\% & 6.55\% & persona\\
\cmidrule{1-7} 
CoCon-TP & 4.45\% & 16.90\% & 33.35\% & 30.95\% & 14.35\% & human \\
	\cmidrule[\heavyrulewidth]{1-7}
	\end{tabular}
\caption{Distribution of persona consistency preferences (\%) for our model (CoCon-TP) compared with seq2seq and persona baselines, according to a five-point Likert scale. A 5 indicates a strong preference for CoCon-TP; a 1 indicates strong preference for the alternative system. }\label{tab:persona_consistency}
\end{table*}

\begin{table*}[t!]
\small
\centering
\begin{tabular}{r r | r| r | r | r | l}
\cmidrule[\heavyrulewidth]{1-7}
 \multicolumn{7}{c}{Distribution of Pairwise Relevance Preferences}\\
\cmidrule[\heavyrulewidth]{1-7} 
Our Method & \multicolumn{1}{c|}{5} & \multicolumn{1}{c|}{4} & \multicolumn{1}{c|}{3} & \multicolumn{1}{c|}{2} & \multicolumn{1}{c|}{1} & Baseline \\ 
\cmidrule[\heavyrulewidth]{1-7} 
CoCon-TP & 13.70\% & 31.05\% & 24.15\% & 22.80\% & 8.30\% & seq2seq\\
CoCon-TP & 10.15\% & 28.60\% & 25.55\% & 25.85\% & 9.85\% & persona\\
\cmidrule{1-7} 
CoCon-TP & 4.75\% & 15.70\% & 28.10\% & 31.95\% & 19.50\% & human \\
	\cmidrule[\heavyrulewidth]{1-7}
	\end{tabular}
\caption{Distribution of relevance preferences (\%) for our model (CoCon-TP) compared with seq2seq and persona baselines, according to a five-point Likert scale. A 5 indicates a strong preference for CoCon-TP; 1 indicates strong preference for the alternative system. }\label{tab:relevance}
\end{table*}

\begin{table*}[t!]
\small
\centering
\begin{tabular}{r r | r| r | r | r | l}
\cmidrule[\heavyrulewidth]{1-7}
 \multicolumn{7}{c}{Distribution of Pairwise Informativeness Preferences}\\
\cmidrule[\heavyrulewidth]{1-7} 
Our Method & \multicolumn{1}{c|}{5} & \multicolumn{1}{c|}{4} & \multicolumn{1}{c|}{3} & \multicolumn{1}{c|}{2} & \multicolumn{1}{c|}{1} & Baseline \\ 
\cmidrule[\heavyrulewidth]{1-7} 
CoCon-TP & 12.85\% & 29.95\% & 27.90\% & 22.60\% & 6.70\% & seq2seq\\
CoCon-TP & 10.45\% & 28.20\% & 30.20\% & 23.55\% & 7.60\% & persona\\
\cmidrule{1-7} 
CoCon-TP & 4.40\% & 15.05\% & 29.80\% & 32.70\% & 18.05\% & human \\
	\cmidrule[\heavyrulewidth]{1-7}
	\end{tabular}
\caption{Distribution of informativeness preferences (\%) for our model (CoCon-TP) compared with seq2seq and persona baselines, according to a five-point Likert scale. A 5 indicates a strong preference for CoCon-TP; 1 indicates strong preference for the alternative system. }\label{tab:informativeness}
\end{table*}

\end{document}